# Bayesian rules and stochastic models for high accuracy prediction of solar radiation

Cyril Voyant[1*], Christophe Darras[2], Marc Muselli[2], Christophe Paoli[2], Marie-Laure Nivet[2] and Philippe Poggi[2]

1-CHD Castelluccio, radiophysics unit, B.P85 20177 Ajaccio- France

2-University of Corsica/CNRS UMR SPE 6134, Campus Grimaldi, 20250 Corte – France

*corresponding author; tel +33495293666, fax +33495293797, cyril.voyant@ch-castelluccio.fr

Abstract:

It is essential to find solar predictive methods to massively insert renewable energies on the electrical distribution grid. The goal of this study is to find the best methodology allowing predicting with high accuracy the hourly global radiation. The knowledge of this quantity is essential for the grid manager or the private PV producer in order to anticipate fluctuations related to clouds occurrences and to stabilize the injected PV power. In this paper, we test both methodologies: single and hybrid predictors. In the first class, we include the multi-layer perceptron (MLP), auto-regressive and moving average (ARMA), and persistence models. In the second class, we mix these predictors with Bayesian rules to obtain ad-hoc models selections, and Bayesian averages of outputs related to single models. If MLP and ARMA are equivalent (nRMSE close to 40.5% for the both), this hybridization allows a nRMSE gain upper than 14 percentage points compared to the persistence estimation (nRMSE=37% versus 51%).

# 1. Introduction

For 60 years, energy needs are multiplying exponentially to support economic developments, comfort and electricity consumption per capita. The gross inland energy consumption was multiplied by a factor 2 according to the world energy assessment report [1]. Currently, we are at one critical moment of the energy exploitation [2]: we carry out the brittleness and the inconsistency of our way. Indeed, the resources of planet in fossil sediment become exhausted, in addition to the economic consequences; so, it is necessary to find alternatives to the current energy sources [3]. In addition, the use of fossil fuels poses another problem: environmental impacts are massive. Even if it has long been ignored, the preservation of the environment is a global issue, with still significant economic challenges. Enhancing the use of renewable energy is one of the solutions but poses a number of challenges in terms of integration.

Because of their random and intermittent trend, the renewable energies must be integrated on a restricted basis in the electrical distribution grid. This parsimonious insertion is in order to protect it and to warrant quality of supply. In France, this limit was set to 30 % of the instantaneous power by the ministerial order of April 23rd, 2008 [4]. To be able to increase the insertion rate of the renewable energies on the electrical distribution grid, solutions are studied and applied. In France for example, the CRE (French Energy Regulation Commission) studies the means to control the fluctuations in these intermittent energies [3]. The CRE is a French independent authority (created on March 24th, 2000) managing industrials tenders related to grid integration of "fatal" energy sources. The solutions proposed by the authority are stipulated in the call for tender of the PV energy and the WT energy [3]. These solutions consist globally in coupling the renewable energies with a storage method (hydrogen, batteries, etc.). However, this coupling is not sufficient if the management of the storage is not mastered. So it is essential to be able to anticipate the renewable energies production. The issue of this paper is the global radiation prediction in order to effectively manage the storage. The association of storage and solar predictive methods allow to guarantee an available energy for the electrical distribution grid [5]. The storage absorbs the strong fluctuations and the surpluses of power and it shall fill the defect of power (we have a PV output power fluctuations smoothing). The manager of the electrical grid could thus estimate the available production of the next day of this power plant [6]; he could so, better manage the supply and demand adequacy. The most interesting prediction horizon for the grid manager is noted h+24 representing 24 predictions computed before 18:00 and covering hour-by-hour, the global solar radiation profile of the next day [7]. The goals of this work could be to integrate these predictions tools to a project like the MYRTE platform [3]. So we will know if an energy production system coupling a PV array and storage (electrolyzer, H2 and O2 tanks and fuel cell in our case) associated with a prediction tool, allows to be seen by the electrical distribution grid manager as a reliable energy supply. In this case, load profiles would be created from forecasted meteorological data, and so the storage/destocking would be also created one day ahead. More the state of tank is wrong and more the daily power supply failures are important. In this kind of platform, the storage doesn't support the PV but the grid by controlling the injected energy into the electrical network.

The global radiation forecasting is the name given to the process used to predict the amount of solar energy available. A lot of predictive methods have been developed by experts around the world. One of the most popular is certainly the numerical weather prediction using mathematical model of the atmosphere to predict the weather based on current weather condition (nRMSE close to 35%) [8]. The second family of models, often called stochastic models, is based on the use of the times series (TS) mathematical formalism [7,9,10]. A TS is described by sets of numbers that measures the status of some activity over time. It is a collection of time ordered observations $x_t$, each one being recorded at a specific time t (period) [11]. A TS model ($\hat{x}_t$) assumes that past patterns will occur in the future. TS prediction or TS forecasting takes an existing series of data $x_{t-k}, .. , x_{t-2}, x_{t-1}$ and forecasts the $x_t$ data values. The goal is to observe or model the existing data series to enable future unknown data values to be forecasted accurately. Thus a prediction $\hat{x}_t$ can be expressed as a function of the recent history of the time series, $\hat{x}_t = f(x_{t-1}, x_{t-2}, ...x_{t-k})$ [12-15]. In preliminary studies [16,17], we have demonstrated that an optimized multi-layer perceptron (MLP) with endogenous inputs made stationary and exogenous inputs (meteorological data) can forecast the global solar radiation time series with acceptable errors (10-20%) [15]. This prediction model has been compared to other prediction methods [16,17] (AutoRegressive and Moving Average called ARMA, k Nearest Neighbor called k-NN, Markov Chains, etc.) and the conclusion was that MLP and ARMA were the best predictors (nRMSE gain close to 2 points) and were similar for the horizon h+1 (prediction one hour ahead, nRMSE close to 15% [16]) and j+1 (prediction one day ahead, nRMSE close to 20% [17]). Moreover, we have shown in previous study [7] that MLP modeling of the global solar irradiation TS can be applied to the

h+24 time horizon prediction (nRMSE lower than 30%). The results demonstrate a higher accuracy with MLP models than with the persistence method. Indeed, the use of MLP to predict the h+24 global radiation horizon is interesting but the chosen architectures, stationarization modes, and the choice of a multivariate analyze, modify greatly the results. Different levels of complexity can be considered, namely: multi-output MLP (with or without exogenous data), MLP committee (with or without exogenous data) or ARMA model.

Considering these findings we propose in this paper new methodology of TS stochastic modeling and of data preprocessing for the prediction of the PV energy (section 2). Then we will present the results of the models mentioned above and expose the performance of prediction in order to cross compare models, and explain the use of Bayesian selection rules (section 3) and then, finally discuss the use of this predictive approach in the case of a real coupling between a PV array and storage.

## 2. Methodologies

Most available global radiation measurements are global horizontal radiation, but it is very rare/uncommon to develop PV stations with no tilted PV modules. To use these historic measures while modeling the global radiation, it is therefore important to be able to tilt horizontal data. The next section will describe the adopted tilt methodology.

### 2.1. The horizontal data problem

Whatever the selected predictive tool, it is necessary in order to develop the majority of the stochastic models to use historical data of global radiation. Depending on the inclination of the PV panels, this type of measures does not often exist on important time intervals. However it is possible to determine the global solar irradiation on tilted plane ($H_{g\beta}$) from horizontal ones. To achieve this, we consider the components of the horizontal global radiation ($H_{gh}$; available on the French Meteorological Organization database), that is to say the direct component called beam ($H_{bh}$), and the diffuse component ($H_{dh}$). According to a past study made on the Mediterranean region [7,14] we decided to use the CLIMED 2 methodology in order to calculate the horizontal diffuse radiation, and the Klucher [18] approach to compute $H_{g\beta}$ (Equation 1).

$$H_{g\beta} = H_{bh}.R_b + R_d.H_{dh} + R \qquad \text{Equation 1}$$

Where $R_b$ is related to the geometric projection (Equation 2, where $\theta$ is the incident angle of the surface, $\varphi$ is the latitude, $\delta$ is the declination, $\omega_h$ is the hour angle, $\beta$ is the tilt angle and $\gamma$ is the solar azimuth angle), $R$ related to the ground scattering (Equation 3 where $\rho$ is the ground albedo) and $R_d$ is related to the Klucher methodology (Equation 4, where $\theta_z$ is the zenith angle).

$$R_b = \frac{\cos(\theta)}{\sin(h)} \qquad \text{Equation 2}$$

$$= \frac{(\sin\varphi \cos\beta - \cos\varphi \sin\beta \cos\gamma)\sin\delta + (\cos\varphi \cos\beta + \sin\varphi \sin\beta \cos\gamma)\cos\delta \cos\omega_h + \cos\delta \sin\beta \sin\gamma \sin\omega_h}{\sin\varphi \sin\delta + \cos\varphi \cos\delta \cos\omega_h}$$

$$R = \frac{1}{2}.\rho.H_{gh}.(1 - \cos(\beta)) \qquad \text{Equation 3}$$

$$R_d = \frac{1}{2}.(1 + \cos\left(\frac{\beta}{2}\right)).(1 + F.\sin^3\left(\frac{\beta}{2}\right)).(1 + F.\cos^2(\theta).\sin^3(\theta_z))$$

where $F = 1 - \left(\frac{H_{dh}}{H_{gh}}\right)^2 \qquad \text{Equation 4}$

The various parameters involved in these equations are classical and are related to the celestial mechanics. The CLIMED2/Klutcher methodology give us Equation 5 with clearness index defined by $k_t = \frac{H_{gh}}{H_0}$, and scattered ratio by $f = \frac{H_{dh}}{H_{gh}}$).

$$\left. \begin{array}{l} f = 0{,}995 - 0{,}081 k_t \quad \text{for } k_t \leq 0{,}21 \\ \\ f = 0{,}724 + 2{,}738 k_t - 8{,}32 k_t^2 + 4{,}967 k_t^3 \quad \text{for } 0{,}21 < k_t < 0{,}76 \end{array} \right\} \quad \text{Equation 5}$$

$$f = 0{,}18 \quad for \ k_t > 0{,}76$$

(Equation 5, clearness index defined by $k_t = \frac{H_{gh}}{H_0}$, and scattered ratio by $f = \frac{H_{dh}}{H_{gh}}$).

On the Mediterranean region, the CLIMED2/Klucher methodology [18], which is described by the Equation 5 gives very good result (clearness index defined by $k_t = \frac{H_{gh}}{H_0}$, and scattered ratio by $f = \frac{H_{dh}}{H_{gh}}$).

The next section details the dedicated estimator to model the global radiation time series.

### 2.2. Major predictive models

Most models used in time series prediction as artificial neural networks (ANN) or ARMA are stationary models, the operating condition is therefore dependent on the stationarity of the data to predict [19]. Global radiation as almost all of the meteorological phenomena has not a stationary nature. It is therefore appropriate to transform the input solar radiation data into acceptable input from a modeling point of view, namely in stationary form. There are many methodologies to make these input stationary such as the clearness index ($k_t$), or Clear Sky Index (CSI). Following previous work we decided to use the Clear Sky Index [14]. The global radiation clear sky model (CS) selected to construct the CSI is the SOLIS model of Mueller [20] described by Equation 6 (*h* is the solar elevation, $\tau_g$ is the global optical depths and $H_0'$ the modified extraterrestrial irradiance).

$$CS = H_0' \cdot exp\left(-\frac{\tau_g}{\sin^g(h)}\right) \cdot \sin(h) \qquad \text{Equation 6}$$

CSI is defined by the ratio between measured global radiation and computed clear sky radiation like it is shows in Equation 7 ($H_{g\beta}$ is the tilted global radiation, and $CS_\beta$ is a tilted version of *CS* computed with the Klucher methodology).

$$CSI = H_{g\beta}/CS_\beta \qquad \text{Equation 7}$$

Note that all hours of the day are not interesting for the PV production, thus we have selected the hours where the radiation is potentially the most important (8:00-16:00, nine hours in true solar time). The errors estimation doesn't take into account the data outside this interval (errors values would be lower but without physical meaning). The data used during this study are the hourly horizontal global radiation (Wh.m$^{-2}$) from 1998 to 2011 on the site of Ajaccio (Corsica, France, 42°09'N and 9°05'E, 38 meters), providing by Météo-France (the French national meteorological organization; pyranometer CM 11 Kipp & Zonen with a sensibility of 6μV/W.m$^{-2}$). Less than 3% of missing data was replaced by the average of global radiation for the concerned hours and days. Next are described the MLP and ARMA approaches as the hybrid methodology based on Bayesian inferences. Note that the persistence model used here as a naïve reference model is defined by a forecast equal to the last known value of the time series (that is to say from time *t* to time *t+1*).

### 2.2.1. Artificial neural Networks

Although many ANN architectures exist, Multi-Layer Perceptron (MLP) remains the most effective and popular [16]. An MLP is made of several layers: one input layer, one or several intermediate layers and one output layer [21]. In the h+24 case, the main characteristic of the MLP concerns the global architecture of the neural network. In a previous study [13] we have discussed three different types of configuration for the MLP organization:

- MLP committee that is to say, nine MLP models each of them dedicated to one given hour (considering that cloud occurrences and mist appear often at the same hours);

- Multi-output MLP: a nine-output MLP model (one output related to each hour of prediction assuming that global solar irradiation for each hour is related to previous measurements);

- A one-output MLP model used nine times (each prediction is considered chronologically like inputs of the network).

This previous study [7] encourages to consider the methodology based on the multi-output MLP (one output related to each hour of prediction). It gave the best result. This approach is a conventional methodology based on the sliding window principle. Measures of global irradiation are chronologically ordered to build up the MLP input vector (Figure 1 ; 27 inputs correspond to measurements related to 3 days of 9 hours each [7]).

In this study, MLP has been compiled with the Matlab© software and its Neural Network toolbox. The main characteristics chosen and related to previous works [10,14,16] are: the hyperbolic tangent (hidden) and identity (output) activation functions, the Levenberg-Marquardt learning algorithm with a max fail parameter before stopping training set to 3 (early stopping method limiting the overtraining). In this study, training, validation and testing data sets were respectively set to 80%, 20% and 0% (Matlab© parameters) and inputs are normalized between − 0.9 and + 0.9. The training phases concern the 12 first years of the global solar irradiation values covered in the set of data (1998-2009). The testing phase for prediction results uses the remaining last two years of the data set (2010-2011).

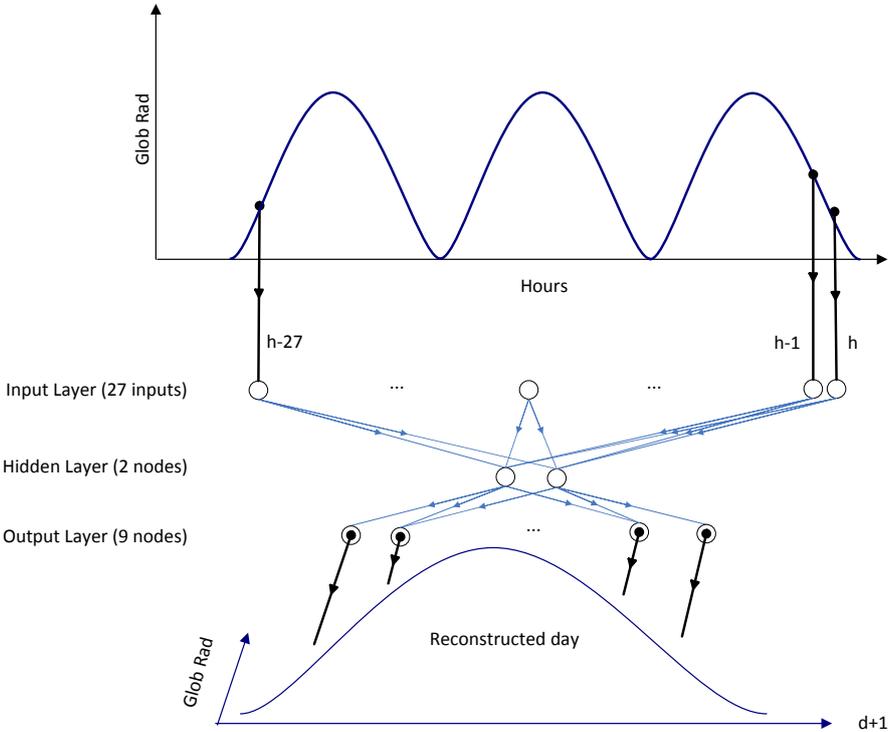

Figure 1: MLP methodology

In addition to the global radiation inputs, it makes sense with the MLP to also use variables of different nature (called exogenous) in order to increase the quality of prediction. The choice of these variables is not trivial, and it is essential to use a robust and objective variables selection methodology. Often the correlation coefficient is used but it indicates only the linear dependence between the variables. The MLP is dedicated to the non-linear estimation, the use of correlation coefficient or partial correlation coefficient induces a bias in the methodology. An alternative is possible using the mutual information (*MI(X,Y)*, equation 8) [22] measuring the mutual dependence of both variables. Mutual information can be expressed as a combination (Equation 8) of marginal and conditional entropies (respectively $H(X)$ and $H(X|Y)$).

$$MI(X,Y) = H(X) - H(X|Y) \qquad \text{Equation 8}$$

To work with more useful quantities like joint probability distribution function ($p(x,y)$) and marginal probabilities ($p(x)$ and $p(y)$), it is preferable to use the Equation 9, or the relative mutual information (*rMI*, Equation 10) resulting of the normalization by the maximum *MI*.

$$MI(X,Y) = \sum_y \sum_x p(x,y) \log\left(\frac{p(x,y)}{p(x)p(y)}\right) \qquad \text{Equation 9}$$

$$rMI(X,Y) = \frac{MI(X,Y)}{MI(X,X)} = \frac{MI(X,Y)}{H(X)} \qquad \text{Equation 10}$$

### 2.2.2. The Autoregressive moving-average

For ARMA models, the methodology is slightly different from the previous case because it is not possible to generate multi-output models. The chosen approach [7] corresponds to the use of nine independent ARMA models (one pattern per hour, Figure 2).

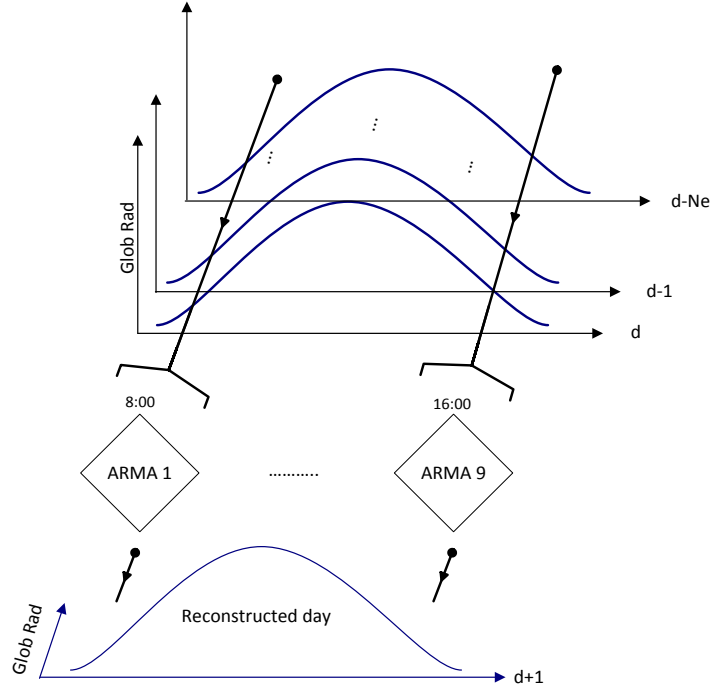

Figure 2: ARMA methodology

Each ARMA model is based on two elementary models: the MA model and the AR model. They are defined like a regression on the last residues and the last measures [23]. A more general so-called ARIMA(p,d,q) model (dedicated to difference-stationary time series) is built from AR(p), MA(q) models and $d^{th}$ difference of the series. This model is defined by the Equation 11 (*L* is the lag operator, $i \in [1, p]\ and\ j \in [1, q]$). The ARMA model (the most popular for TS prediction 13]) is a particular case of ARIMA with *d* equal to 0.

$$\left(1 - \sum_i \phi_i L^i\right)(1-L)^d X_t = \left(1 - \sum_j \theta_j L^j\right)\epsilon_t \qquad \text{Equation 11}$$

Previous study concerning the h+24 horizon shows that for optimal result, only AR component is necessary (AR(1) for all models [7,14]).

### 2.3. Bayesian decision rules for hybrid models

It is interesting to develop tools which allow us to critically judge the available models. Experience shows that there is rarely a model that is always very good (depending on seasons, locations, etc.). In this context, the hybridization or the model selection related to the current observations seems to be the solution: getting the best performance of each model when it is more efficient. In this logic, the goal is to find rules that allow in a given context, to choose the best model. One method (among many, like data mining, classical statistic, etc. [24]) consists to use the Bayes inferences. In the Bayesian context, a probability P represents a degree-of-belief [25] based on known information. The problem consists in inferring which model is most plausible given the data. With hypothesis (hyp) and observation (obs) concept, the Bayes theorem can be expressed by P(hyp|obs) = P(obs|hyp).P(hyp)/P(obs). This formula allows updating our degree of confidence, taking into account the observations. Applying the Bayes theorem on the data set $D$ (observation) and the $j$ methods of hypothesis of prediction called ($1 < i < j$; $M_i$ is the ith methods), we get the Equation 12.

$$P(M_i|D) = \frac{P(D|M_i).P(M_i)}{\sum_j P(D|M_j).P(M_j)} \quad \text{Equation 12}$$

$P(D|M_i)$ is the marginal likelihood or evidence for model $M_i$ and defined from the parameters $\theta_i$ (like shown in the Equation 13) and the quantity $P(M_i)$ represents a prior belief for model $M_i$ [9,26,27].

$$P(D|M_i).P(M_i) = P(M_i).\int P(D|\theta_i, M_i).P(\theta_i|M_i)d\theta_i \quad \text{Equation 13}$$

A simple rule of selection between models is related to the comparison of the $P(M_i|D)$ factors. If $P(M_i|D) > P(M_j|D)$ then the $M_i$ will model the data more closely than $M_j$. Like the denominator of the posterior probability of each model $P(M_i|D)$ does not depend on the model, it is only the numerator that allows deciding between models. To compute the numerator, it is possible to consider $n$ series of real and measured data (noted $E_n$) and not the parameter $\theta_i$. With the hypothesis of variables independence, the equation 13 is updated and the selection parameter is detailed in the Equation 14.

$$P(D|M_i).P(M_i) = \prod_n P(E_n|M_i).P(M_i) = P_i \quad \text{Equation 14}$$

In the global radiation time series forecasting, the selection rule becomes the Equation 15.

$$if\ P_i > P_k\ then\ \hat{x}(t+1) = \hat{x}^i(t+1)\ else\ \hat{x}(t+1) = \hat{x}^k(t+1) \qquad \text{Equation 15}$$

It is then possible to describe a Bayesian average of the available predictor, like shown in the Equation 16.

$$\hat{x}(t+1) = \frac{\sum_i P_i \hat{x}^i(t+1))}{\sum_i P_i(t+1))} \qquad \text{Equation 16}$$

By referring to previous studies [14], four modified endogenous and exogenous data are tested ($E_n$ with $n \in [1,4]$); the mean global radiation of the actual day (MGR), the daily index of the actual day (DI; between 1 and 365), the differential pressure between the actual day and the day before (Pdt), and the mean error of prediction computed previous day (PME). In theory, the parameter $P(E_n|M_i)$ is equal to $P(E_n(1) \dots E_n(t-1), E_n(t)|M_i)$ but $\forall n$ and if $t_0(\in \mathbb{N}) > 1$ then $rMI(x(t+1), E_n(t-t_0)) \to 0$. For simplicity, only the first lag is considered. Parameters related to the Bayesian inferences are computed during the learning step and reported during the test phase.

## 3. Results

In this section, the results related to the data preprocessing will be shown (tilting and clear sky index generation). Then using these preprocessed data, the global radiation prediction with conventional models (ARMA, MLP and persistence) will be exposed. Finally, the hybridization of the models using Bayesian selection rules will finish the results section.

### 3.1. Data preprocessing

To study and estimated the tilted global radiation with stochastic models, it is necessary to generate large learning time series from horizontal measurement. The inclination process is decisive in the prediction workflow. During this section, the considered error estimation is the nRMSE (normalized Root Mean Square Error in % and defined by $\sqrt{<(x-\hat{x})^2>/<x^2>}$).

### 3.1.1. Inclination of the data

The validation of the model presented in the section 2.1 is done with measurements cross-comparison. Measures are related to horizontal (0°) and tilted global radiation (45° and 60°) during one month (June 2010). For more information, interested reader can read the paper of Notton et al. [18] which describes an exhaustive benchmark of inclination modeling. The used methodology (CLIMED2 and Klucher) allows generating new times series from horizontal measurements. Global horizontal radiation is used to estimate 45° and 60° global radiation. The computed nRMSE are close to 11% for 45° and 23% for 60°. More the tilt angle is important, bigger is the error. Note that in the 30° case the error is certainly lower than 10%. Used methodologies are relevant and usable for other tilt angles like shown in previous study [14]. In the next, data used will be 30°-tilted (corresponding to the maximum annual PV energy for the Ajaccio latitude).

### 3.1.2. Clear sky index

The CS generation is necessary to make the global time series stationary. Indeed, the CSI (index used during the MLP and ARMA modeling) requires the global radiation without cloud occurrences like described in the section 2.2. The Figure 3 shows an example of the impact of the pretreatment during 200 consecutive and sunny days for 30°-tilted irradiation (the nights are not considered ; 9 hours a day between 8:00 and 16:00). With a good stationary process, it is possible to model with high accuracy the global radiation, to consider only a forecasted CSI equal to 1 (e.g. global radiation prediction equal to clear sky computing). Stochastic models are only necessary to take into account the "abnormal points" induced by the clouds attenuation.

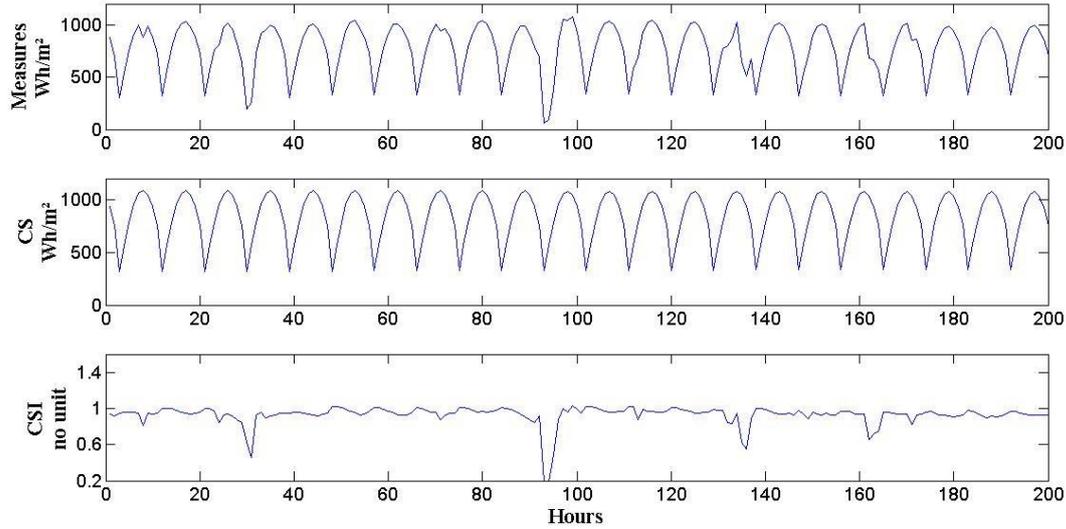

Figure 3: Pretreatment of the global radiation time series

### 3.2. Predictions with basic models

Models studied in this section are MLP, ARMA, CS (computed global radiation without taking into account cloud cover) and persistence. The ARMA models use the methodology presented in section 2.2.2 of 9 patterns each one dedicated to one specific hour. In fact, the best configuration (after CSI transformation) for all the patterns is AR(1). For MLP, the model is based on a 9-outputs network, with 27 endogenous nodes and 2 hidden neurons (1 layer). The pressure is intuitively a parameter allowing anticipating the cloud occurrences. Six sub-variables are constructed from pressure time series, the minimum pressure (*Pm*), the maximum pressure (*PM*) and the mean pressure of the current day (*Pa*), the daily trend pressure (*Pdt*, difference between the daily pressure of current day and the day before), the pressure measured at 18:00 and 12:00 the current day (*P18* and *P12*), and the intra-day variation (*Pid = P18-P9*). Concerning the selection of exogenous inputs with the relative Mutual Information method (*rMI*), the Figure 4a gives result of the mutual dependence between the 12:00 global radiation time series of the day *d+1* and the computed series related to the pressure of the day before.

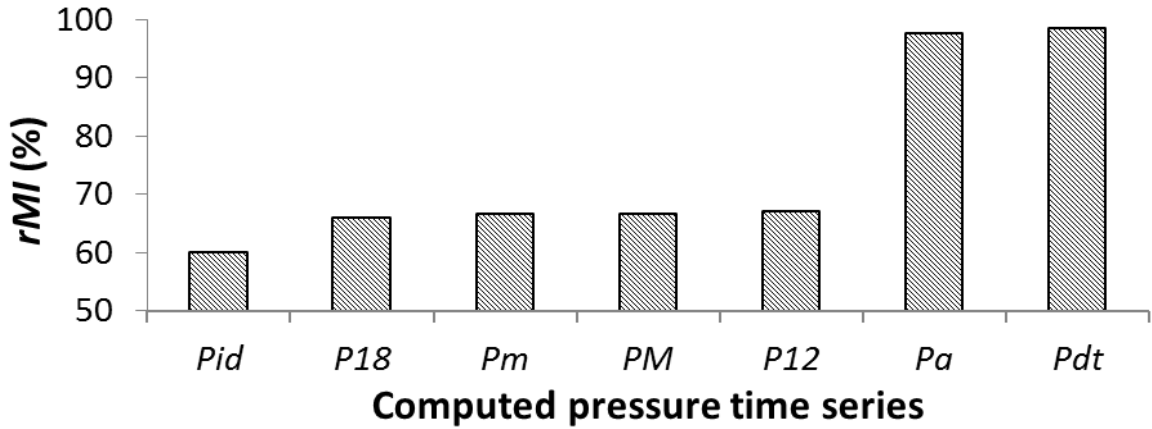

Figure 4a: *rMI* between the 12:00 global radiation time series at day *d+1* and the 7 series related to the pressure at day *d*

Two variables (*Pa and Pdt*) are more dependent on the global radiation, according to the parsimony principle, it seems interesting to construct a test in order to select only the more interesting of these both variables in term of statistical relationship. The Figure 4b represents the difference between the *rMI* of *Pa* and *Pdt* computed from the nine hours of the next day. A positive value means that *Pdt* is more linked than *Pa*.

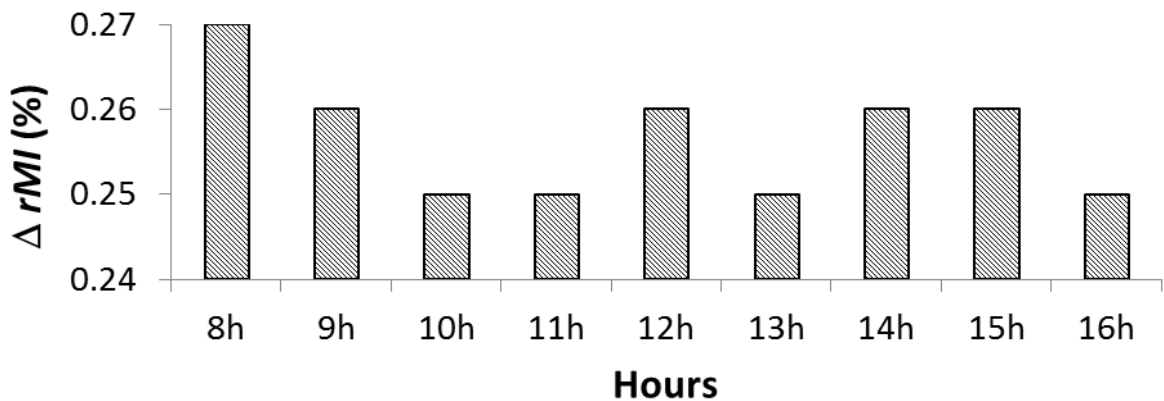

Figure 4b: Difference between the *rMI* of *Pa* and *Pdt* (day *d*) computed from the nine hours of the next day (*d+1*)

As we can see, for all hours Δ *rMI* is positive. Considering this fact, in the next steps, one exogenous input based on *Pdt* (difference between the mean pressure at day *d* and *d-1* to predict the irradiation at day *d+1*) is added in the first layer of the MLP. The four models studied can be classified into two groups, the stochastic models (MLP and ARMA) and the naive models (CS and persistence). The first group gives the best results concerning the 24 hours ahead prediction of the two years considered (2010-2011). For ARMA, the nRMSE is 40.32% and for MLP it is 40.55% (± 0.2% depending on the random initializations of weights and bias [24]). The two other models give less interesting results, for persistence and CS nRMSE are respectively 50.62% and 53.77%. If the models of the first group are equivalent (± 0.2 point), for the second, the persistence model is widely better than the CS estimation. In the next of this document CS estimation will no longer be used. Before giving results related to the Bayes inferences, it is necessary to understand why, for this horizon, even the better model gives nRMSE upper than 40% (note that in the h+1 horizon the nRMSE was closed to 15% [19]). The Figure 5 represents the measured and estimated global radiation profiles (better model, ARMA) for three characteristic hours of the day (beginning of day, midday and end of day). Note that the unit is Wh.m$^{-2}$ referring to the radiation during an hour interval, i.e. the 8:00 radiation value corresponds to the period between 7:00 and 8:00.

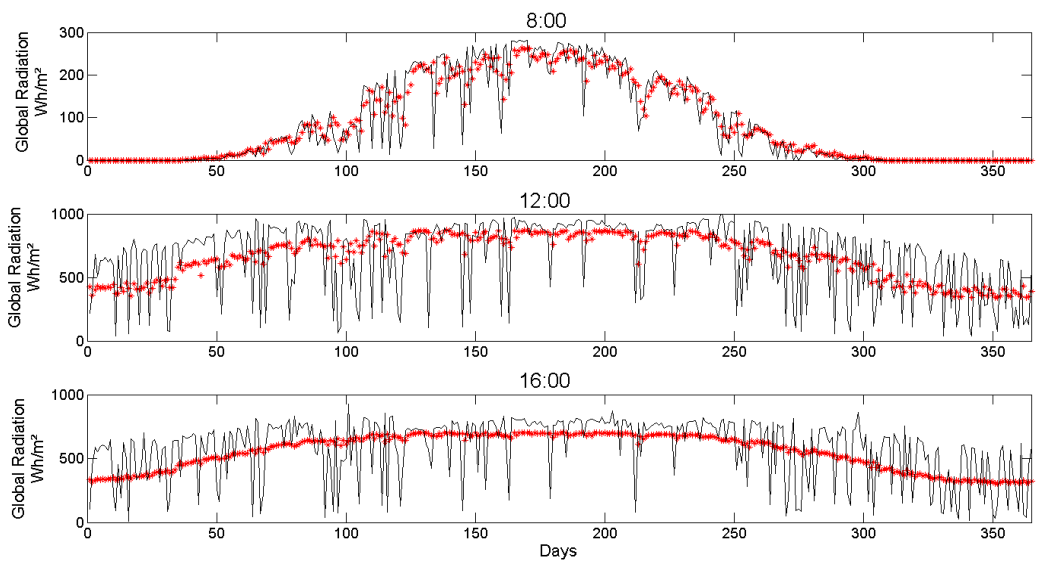

Figure 5: Measured (lines) end estimated (with ARMA model; marks) global radiation respectively for 8:00, 12:00 and 16:00

If for 8:00 profile, the model fits correctly the measures, concerning the two other hours, high frequencies of the signal are not taken into account. The model proposed is limited to an average or a smooth of the measured signal. The prediction does not seem powerful. This phenomenon is also observed in the Figure 6, which represents the classical hourly profile of the global radiation. At the 13$^{th}$ day (represented by an arrow), the model is totally misguided. It forecasts a classical global radiation day while the measures do not exceed 600 Wh/m². In the next part, we will see if the Bayesian rules allow to overcome this problem and to reduce the prediction error.

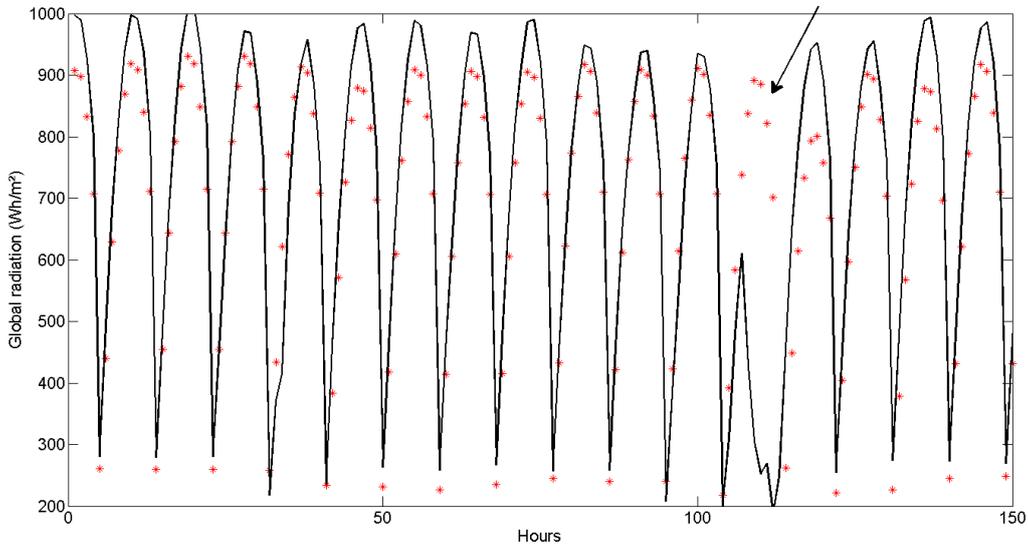

Figure 6: Measured (lines) end estimated (with ARMA model; marks) global radiation for 17 days (May 2011)

### 3.3. Bayesian selection

In the 2.3 section, we have seen that it is possible to mix or to hybrid the forecasting methodologies. The two proposed approaches are based on Bayesian selection and Bayesian average of models. To use these approaches, it is first necessary to explicit the $P_i$ expression for the three studied models (MLP, ARMA and persistence). $P_i$ represents the probability that the model $i$ will be the best estimator given the class of the studied parameter (mean global radiation (MGR), daily increment (DI) or differential pressure (Pdt)). In all case, the sum of the probability $P_{MLP}+P_{ARMA}+P_{persistence}$ is 1.

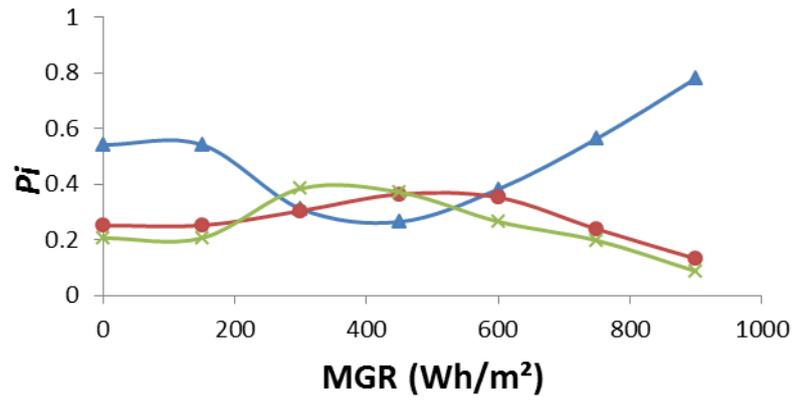

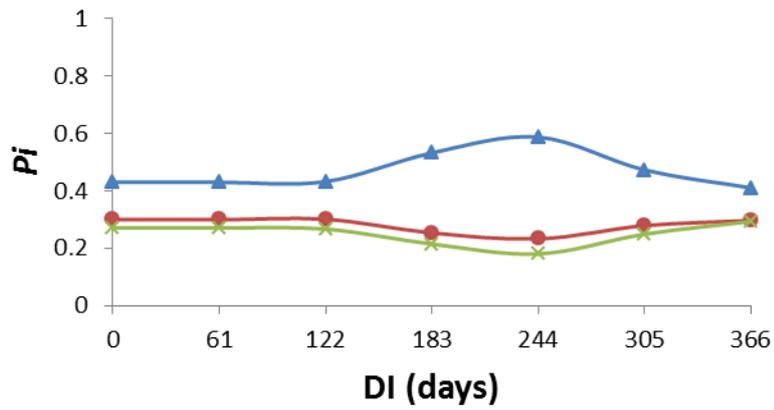

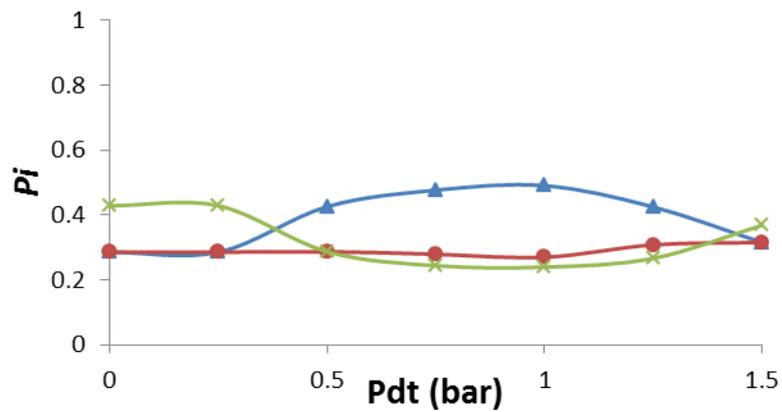

Figure 7: $P_i$ expression for mean global radiation (MGR), daily increment (DI) and Pressure (Pdt) related to Persistence (▲), ARMA (●) and MLP (✕) models

Due to the Figure 7, we note that DI parameter cannot be used in the context of model selection or the Bayesian average. The $P_i$ curves do not intersect showing that persistence is always the best predictor. In the next, it will not be used. The last parameter studied is related to the previous mean error (PME). This construction is particular and do not correspond to the MGR, DI and Pdt methodologies (not compatible). It is only applicable for the selection of model (two by two selections) and not for the Bayesian average. Considering two models $i$ and $j$, and considering the previous mean error of the lag t ($PME^{i,j} = \langle \hat{x}^{i,j}(t) - x^{i,j}(t) \rangle$), it is possible to construct the following difference $\Delta PME = PME^i - PME^j$. Trends of $P_i$ and $P_j$ correspond to a Heaviside step function ($H$), so $P_i = H(-\Delta PME)$ and $P_j + P_i = 1$. In this section ten models are compared, the three previous based on a single predictor (persistence, ARMA, MLP) and the following seven hybrid predictors based on Bayesian rules:

-hybrid 1: ARMA, MLP and persistence selection according to the PME rule

-hybrid 2: ARMA, MLP and persistence selection according to the Pdt and MGR rules

-hybrid 3: ARMA, MLP and persistence selection according to the MGR rule

-hybrid 4: ARMA, MLP and persistence selection according to the Pdt rule

-hybrid 5: ARMA, MLP and persistence average according to the Pdt and MGR rules

-hybrid 6: ARMA, MLP and persistence average according to the MGR rule

-hybrid 7: ARMA, MLP and persistence average according to the Pdt rule

Note that other hybrid models can be generated, we have decided to limit this study to these seven models because we believe they are the most interesting (Pdt and MGR are equivalent and interchangeable variables and PME, by construction is different and is not compatible with "average" but only with "selection" mode). Table 1 draws conclusions of comparisons. We consider the error (nRMSE) of prediction step by step and the cumulative error of the prediction (hourly predictions integration during 24 hours of predictions and measures). If $x_i$ is the hourly and $x_{cum}$ the daily value of the global radiation (9 hours a day) $x_{cum} = \sum_{i \in [1,9]} x_i$), the daily prediction error becomes $\sqrt{\langle (x_{cum} - \hat{x}_{cum})^2 \rangle / \langle x_{cum}^2 \rangle}$.

| Models | Prediction error (nRMSE; %) | Daily prediction error (nRMSE; %) |
|---|---|---|
| **MLP** | 40.55 | 31.63 |
| **ARMA** | 40.32 | 32.49 |
| **Persistence** | 50.62 | 39.94 |
| **Hybrid 1** | **36.59** | **28.39** |
| **Hybrid 2** | 50.32 | 39.69 |
| **Hybrid 3** | 45.97 | 36.29 |
| **Hybrid 4** | 50.13 | 39.53 |
| **Hybrid 5** | 77.19 | 70.39 |
| **Hybrid 6** | 41.91 | 33.67 |
| **Hybrid 7** | 42.24 | 33.81 |

Table 1: Hourly prediction error and cumulative prediction error (measures and predictions are integrated during 24 hours). Best results in bold.

The best model shown by this Table is the model "Hybrid 1: ARMA, MLP and persistence selection according the PME rule", it allows a reduction by about 4 points according to the best single predictor MLP, this gain is important and this method is relatively easy to implement (nRMSE=36.6% for Hybrid 1 and 40.6% for simple MLP). More the prediction is precise and more the management of storage/redistribution is easy for the PV station manager. The global radiation estimation of the next day is directly linked with the load profile shape of the next days, so the owner of the PV station capable of anticipate the first will able to propose to the grid manager an gainful solution: installation sustainability (without penalties and with profitability) and electrical grid stability contribution. Concerning the daily prediction error, the gain is less important but upper than 3 points. Concerning the other hybrid models, we see that using the Bayes inferences is not systematically interesting. The hybrid model 5 (ARMA, MLP and persistence average according the Pdt and MGR rules) is not only worse than simple stochastic models (nRMSE > 70% for prediction and daily prediction error), but also not better than naïve model (persistence; prediction error equal to 50.6% and daily prediction error equal to 39.9%).

## 4. Conclusion

In this paper, we have proposed a prediction methodology relevant for the PV station management. The horizontal measured data were tilted in order to simulate a concrete case (tilt_angle = latitude = 30°). We have shown that the h+24 horizon is compatible with the MLP and ARMA forecasters (nRMSE close to 40.5% for the both). Exogenous data are chosen by mutual information method and added in the input layer of the MLP. This methodology is not often used to the detriment of correlation coefficient whereas the linearity of the meteorological time series is not shown. The proposed MLP optimization is one of the strengths of this paper. Moreover, actually, there is no consensus regarding to global radiation prediction 24 hours ahead. Numerical weather predictions are often an alternative but the error generated is important (nRMSE ~ 40% for central Europe [23]), and we imagine that in the Corsican case, the orography and the small size of the island is not consistent with the satellite acquisition and the related resolution. The error is certainly not lower than 40%. In this paper, concerning h+24 predictions, we have too demonstrated that MLP and ARMA are equivalent but widely better than persistence (nRMSE gain close to 10 percentage points). In the daily integration case, MLP gives better results than ARMA (nRMSE=31.6% versus 32.5%). The hybridization of these three predictors is difficult, even if the model 1 - ARMA, MLP and persistence selection according the PME rule - designed in this paper induces very good results (nRMSE=36.6%), we note that for other hybrid models, the generalization is not allowed. The global methodology of this paper is detailed in the Figure 8. The integration of the PV energy needs studies like this, the next step will be to test this approach on a concrete case of PV installation coupled with a storage system (like the MYRTE platform [3] detailed in introduction). Moreover, one way of improvement is also to combine the hybrid model described here and a classical and sophisticated numerical weather prediction like the AROME modeling system from Météo-France. If the first takes into account only the intrinsic characteristics of the global radiation time series (stochastic models), the second takes into account all the meteorological quantities and estimate the weather from the evolution of the atmosphere over a few day. The combination of these two approaches should allow proposing a successful prediction tool.

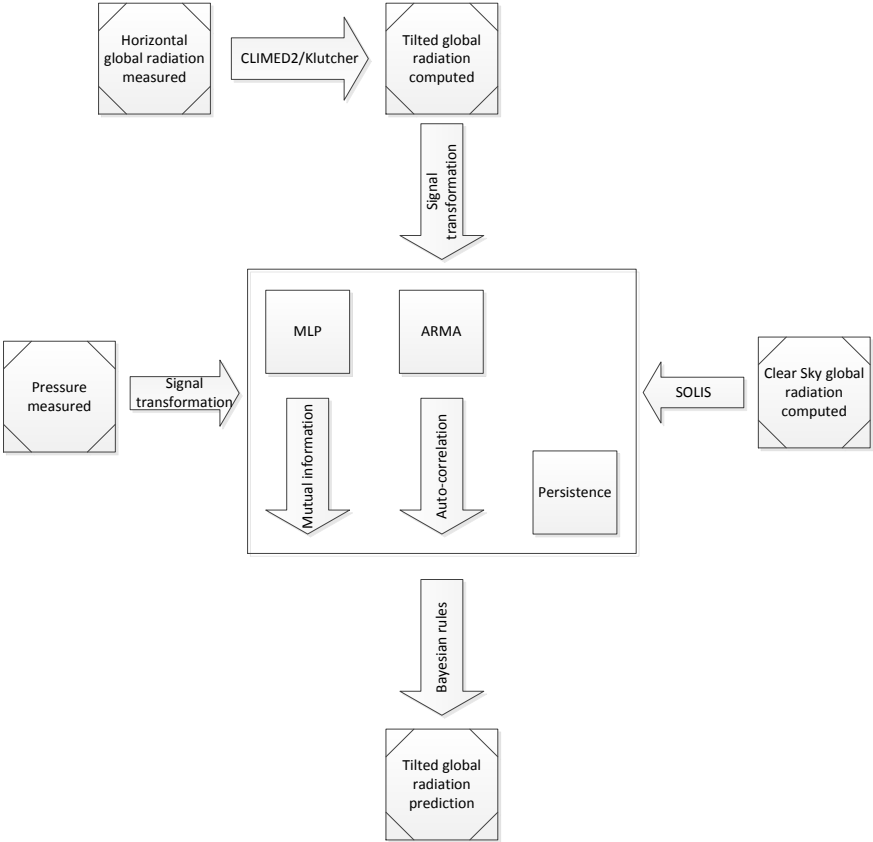

Figure 8: Modeling methodology

# 5. References

[1] http://webarchive.iiasa.ac.at/Research/TNT/WEB/Publications/publications.html

[2] Carl-Jochen W. Hydrogen energy Abundant, efficient, clean: A debate over the energy-system-of-change. International Journal of Hydrogen Energy 2009;34(14):S1–52.

[3] Darras C, Muselli M, Poggi P, Voyant C, Hoguet JC, Montignac F. PV output power fluctuations smoothing: The MYRTE platform experience. International Journal of hydrogen energy 2012;37(19):14015-14025.

[4] http://www.legifrance.gouv.fr/affichTexte.do?cidTexte=JORFTEXT000018697930

[5] Aki H, Taniguchi Y, Tamura I, Kegasa A, Hayakawa H, Ishikawa Y, et al. Fuel cells and energy networks of electricity, heat, and hydrogen: A demonstration in hydrogen-fueled apartments. International Journal of Hydrogen Energy 2012;37(2):1204–13.

[6] Mayer D, Wald L, Poissant Y, Pelland S. Performance prediction of grid-connected photovoltaic systems using remote sensing. Report IEA-PVPS T2-07; 2008:18.

[7] Voyant C, Randimbivololona P, Nivet ML, Paoli C, Muselli M. 24-hours ahead global irradiation forecasting using Multi-Layer Perceptron. Meteorological applications 2013, In press.

[8] Lorenz E, Hurka J, Heinemann D, Beyer HG. Irradiance Forecasting for the Power Prediction of Grid-Connected Photovoltaic Systems. IEEE Journal of Selected Topics in Applied Earth Observations and Remote Sensing 2010;2:2–10.

[9] Yacef R, Benghanem M, Mellit A. Prediction of daily global solar irradiation data using Bayesian neural networks: a comparative study. Renew Energy 2012;48:146–54.

[10] Sfetsos A, Coonick AH. Univariate and multivariate forecasting of hourly solar radiation with artificial intelligence techniques. Solar Energy 2000;68(2):169-178.

[11] Brabec M, Badescu V, Paulescu M. Nowcasting sunshine number using logistic modeling - Meteorology and Atmospheric. 2012 – Springer.

[12] Benghanem M, Mellit A, Alamri SN. ANN-based modelling and estimation of daily global solar radiation data: A case study. Energy Conversion and Management 2009;50(7):1644–55.

[13] Mellit A, Kalogirou SA. Artificial intelligence techniques for photovoltaic. applications: a review. Prog Energy Combust Sci 2008;34:574–632.

[14] Mellit A, Kalogirou SA, Hontoria L, Shaari S. Artificial intelligence techniques for sizing photovoltaic systems: A review. Renewable and Sustainable Energy Reviews 2009;13(2):406-419.

[15] Schlink U, Herbarth O, Richter M, Dorling S, Nunnari G, Cawley G, et al. Statistical models to assess the health effects and to forecast ground-level ozone. Environmental Modelling & Software 2006;21(4):547–58.

[16] Voyant C, Muselli M, Paoli C, Nivet M-L. Numerical weather prediction (NWP) and hybrid ARMA/ANN model to predict global radiation Energy 2012;39(1):341-355.

[17] Paoli C, Cyril V, Marc M, Nivet ML. "Forecasting of preprocessed daily solar radiation time series using neural networks." Solar Energy 2010;84(12):2146-2160.

[18] Notton G, Poggi P, Cristofari C. Predicting hourly solar irradiations on inclined surfaces based on the horizontal measurements: Performances of the association of well-known mathematical models. Energy Conversion and Management 2006;47:1816-1829.

[19] Voyant C, Muselli M, Paoli C, Nivet M. Optimization of an artificial neural network dedicated to the multivariate forecasting of daily global radiation. Energy2011;36(1):348-359.

[20] Mueller RW, Dagestad KF, Ineichen P, Schroedter-Homscheidt M, Cros S, Dumortier D, Kuhlemann R, Olseth JA, Piernavieja G, Reise C, Wald L, Heinemann D .Rethinking satellite-based solar irradiance modelling: The SOLIS clear-sky module. Remote Sensing of Environment 2004;91(2):160-174.

[21] Patra JC. Neural network-based model for dual junction solar cells. Prog Photovoltaic Res Appl 2011;19:33–44.

[22] Kuijper A. Mutual information aspects of scale space images. Pattern Recognition 2004;37(12):2361-2373.

[23] Paulescu M, paulescu E, Gravila P, Badescu B. Weather modeling and forecasting of PV systems operation. Ed. Springer, 2013.

[24] MacKay DJC. A practical Bayesian framework for backpropagation networks. Neural Comput 1992;4:448–72.

[25] Lauret P, Boland J, Ridley B. Bayesian statistical analysis applied to solar radiation modelling. Renewable Energy 2013;49:124-127.

[26] Burden F, Winkler D. Bayesian regularization of neural networks. In: Livingstone DJ, editor. Artificial neural networks: methods and applications. Humana Press; 2008.

[27] Massi Pavan A, Mellit A, De Pieri D, Kalogirou SA. A comparison between BNN and regression polynomial methods for the evaluation of the effect of soiling in large scale photovoltaic plants. Applied Energy 2013;108:392-401.

## 6. Conflicts of interest

None

## 7. List of captions

### 7.1. Table

Table 1: Hourly prediction error and cumulative prediction error (measures and predictions are integrated during 24 hours). Best results in bold.

### 7.2. Figures

Figure 1: MLP methodology

Figure 2: ARMA methodology

Figure 3: Pretreatment of the global radiation time series

Figure 4a: rMI between the 12:00 global radiation time series at day d+1 and the 7 series related to the pressure at day d

Figure 4b: Difference between the rMI of Pa and Pdt (day d) computed from the nine hours of the next day (d+1)

Figure 5: Measured (lines) end estimated (with ARMA model; marks) global radiation respectively for 8:00, 12:00 and 16:00

Figure 6: Measured (lines) end estimated (with ARMA model; marks) global radiation for 17 days (May 2011)

Figure 7: $P_i$ expression for mean global radiation (MGR), daily increment (DI) and Pressure (Pdt) related to Persistence ( 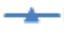 ), ARMA ( 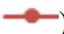 ) and MLP ( 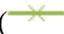 ) models

Figure 8: Modeling methodology